\documentclass[a4paper,twoside]{article}
\usepackage{url}
\usepackage{hyperref} 
\usepackage{placeins}
\usepackage{epsfig}
\usepackage{subcaption}
\usepackage{calc}
\usepackage{amssymb}
\usepackage{amstext}
\usepackage{amsmath}
\usepackage{amsthm}
\usepackage{multicol}
\usepackage{pslatex}
\usepackage{apalike}
\usepackage{algorithm2e}
\usepackage[bottom]{footmisc}
\usepackage{SCITEPRESS}     

\begin{document}

\title{Improve bounding box in Carla Simulator}

\author{\authorname{Mohamad Mofeed Chaar\orcidAuthor{0000-0001-9637-5832}, Jamal Raiyn\orcidAuthor{0000-0002-8609-3935} and Galia Weidl\orcidAuthor{0000-0002-6934-6347}}
\affiliation{Connected urban mobility, Faculty of Engineering,\\ University of Applied Sciences, Aschaffenburg, Germany}
\email{\{MohamadMofeed.Chaar, Jamal.Raiyn, Galia.Weidl\}@th-ab.de}
}


\abstract{The CARLA simulator (Car Learning to Act) serves as a robust platform for testing algorithms and generating datasets in the field of Autonomous Driving (AD). It provides control over various environmental parameters, enabling thorough evaluation. Development bounding boxes are commonly utilized tools in deep learning and play a crucial role in AD applications. The predominant method for data generation in the CARLA Simulator involves identifying and delineating objects of interest, such as vehicles, using bounding boxes. The operation in CARLA entails capturing the coordinates of all objects on the map, which are subsequently aligned with the sensor's coordinate system at the ego vehicle and then enclosed within bounding boxes relative to the ego vehicle's perspective. However, this primary approach encounters challenges associated with object detection and bounding box annotation, such as ghost boxes. Although these procedures are generally effective at detecting vehicles and other objects within their direct line of sight, they may also produce false positives by identifying objects that are obscured by obstructions. We have enhanced the primary approach with the objective of filtering out unwanted boxes. Performance analysis indicates that the improved approach has achieved high accuracy.}

\onecolumn \maketitle \normalsize \setcounter{footnote}{0} \vfill

\section{\uppercase{Introduction}}\label{sec:introduction}
\begin{figure}[h]
    \centering
    \begin{subfigure}{.25\textwidth}
    \centering
    \includegraphics[width=1\linewidth]{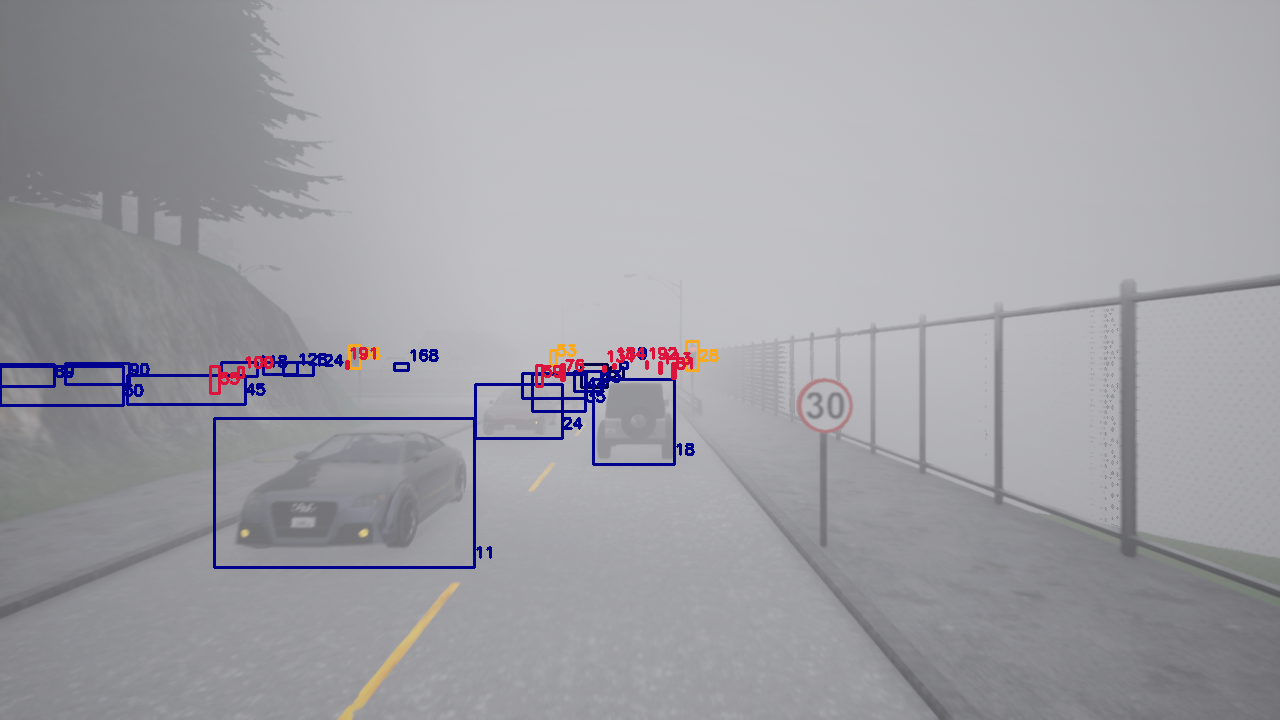}
    \caption{}
    \label{fig:sub1}
    \end{subfigure}%
    \begin{subfigure}{.25\textwidth}
    \centering
    \includegraphics[width=1\linewidth]{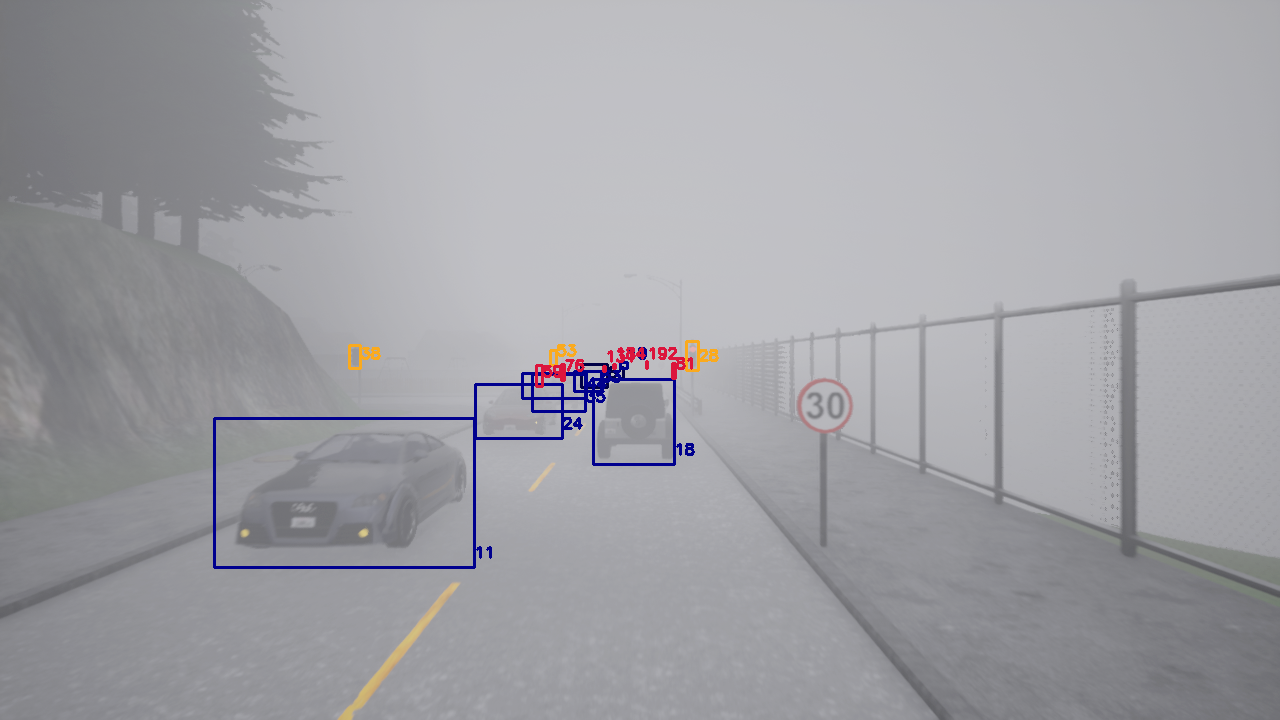}
    \caption{}
    \label{fig:sub2}
    \end{subfigure}
    
    \caption{Comparison between the image before (a) and after (b) the filter.}
    \label{fig:fig01}
\end{figure}

The procedure of bounding box in Carla simulation \cite{Ref01} is implemented by invoking the Bounding Box object. This returns a 3D object box, which we then transform to align with the sensor's coordinate system on the ego vehicle, using vertex transformations as the following formula \cite{Ref02}.\\
\begin{equation}\label{eq1}
    P^{\grave{}}= \begin{pmatrix}
    R & T\\
    0 & 1
    \end{pmatrix}P
\end{equation}
Where \(P=(x,y,z,1)\) is the map of coordinates and \(P^{\grave{}}=(x^{\grave{}},y^{\grave{}},z^{\grave{}},1)\) 
represents the new coordinates with reference point R to the sensor on the vehicle:
\begin{equation}\label{equ2}
    R=\begin{pmatrix}
        cos \theta & -sin \theta & 0\\
        sin \theta & cos \theta & 0\\
        0 & 0 & 1
    \end{pmatrix}
\end{equation}
and Translation \(T\) is 3x1 matrix (or vector) that translates a point in 3D space. Subsequently, we identify all vehicles that fall within the sensor's field of view, accounting for its height and width parameters. Finally, we project the 3D boxes to 2D because we work with images. In Carla, the bounding box accounts for the sensor's coordinates and field of view. However, it does not consider objects that are obscured by other structures, such as buildings. To address this limitation, our methodology involves refining the detection process. We synchronize the sensor data with a Semantic Segmentation camera \cite{Ref01} By conducting a pixel-by-pixel comparison between the sensor and the Semantic Segmentation camera, we are able to filter out unwanted bounding boxes effectively  (Figure\ref{fig:fig01})
\section{\uppercase{RELATED WORKS}}\label{sec:RELATED WORKS}
\begin{figure}
    \centering
    \includegraphics[width=1\linewidth]{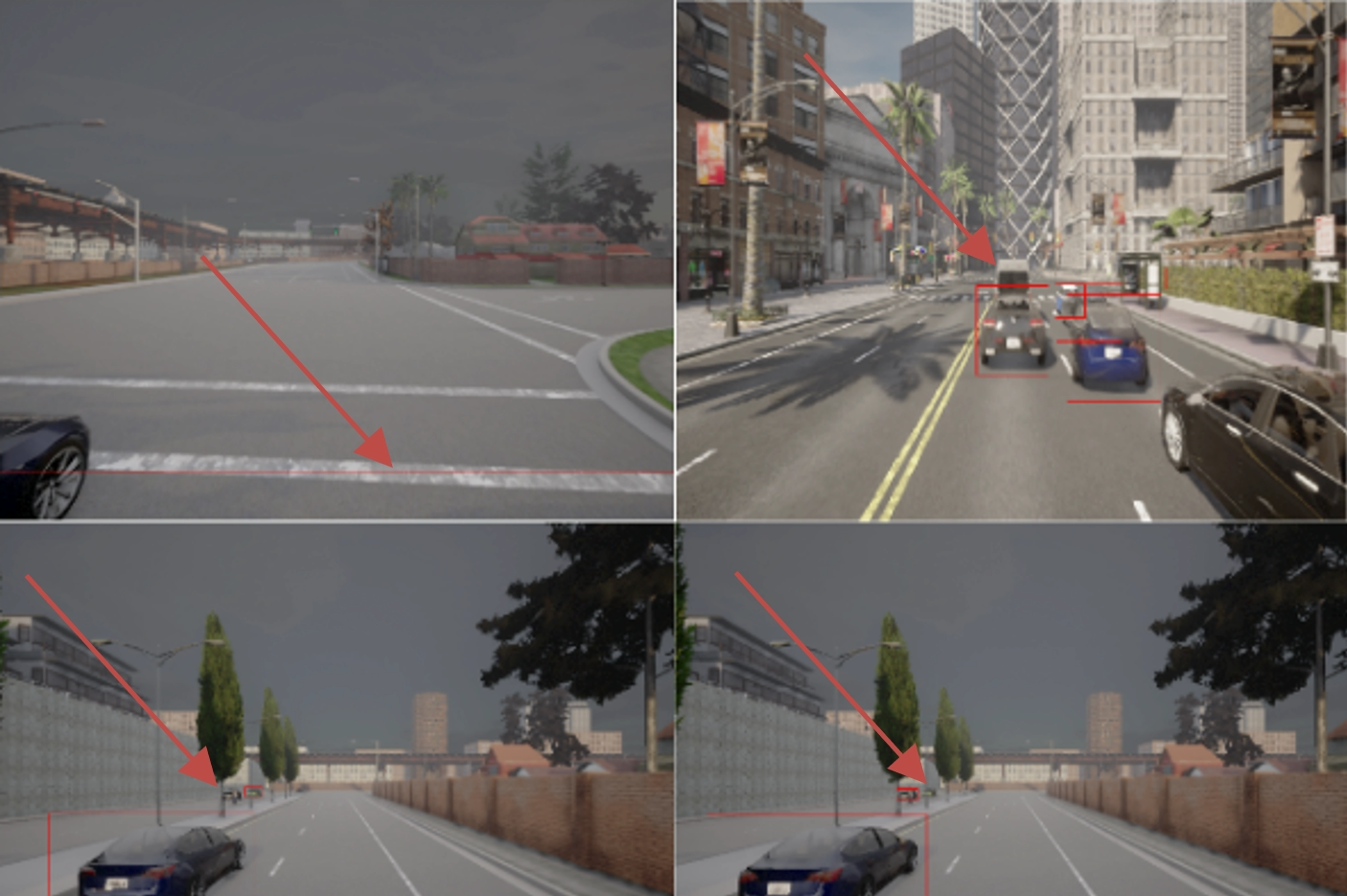}
    \caption{In the work of \cite{Ref04} the filter is using semantic LiDAR. One can see that two objects are filtered while they should be included in bounding boxed (see both down images).   The significant box (only its red bottom line is visible (see Up left image)) was not filtered although it should be. In the filter using depth camera (Up right) one can still see that the object is existing and visible, although it has been wrongly filtered (no bounding box was put around it).}
    \label{fig:fig02}
\end{figure}

The authors \cite{Ref03} Operated the Single Shot Detector (SSD) to train a data generated by the Carla simulator, which was created by capturing 1028 images from RGB sensor with size (640x380) pixels from different maps and different weather conditions. Because of the encountered problems in Carla with the auto generation of a bounding box from the simulator, the authors annotated the labels using Python GUI called labelimg to identify 5 object classes (Vehicles, Bike, Motorbike, Traffic light, and Traffic Sign). The paper provides a valuable contribution to the field of autonomous driving research by demonstrating the effectiveness of using the CARLA simulator to train and test deep learning-based object detection models. On the other hand, \cite{Ref04} generate a new methodology to collect a datasets automatically, which they filtered the objects by taking the semantic LiDAR data and remove all objects which are not in the point cloud of the Lidar sensor. The issue with the Lidar sensor is its lack of precision when dealing with distant objects. The sensor's accuracy diminishes proportionally with the distance to the objects. Additionally, it is highly affected by weather conditions. For instance, when operating in foggy conditions, it negatively impacts the sensor's accuracy. Therefore, we have implemented flexible criteria that align with this requirement. Further elaboration on these points will be presented in this paper. The work of \cite{Ref05} utilized the depth camera to remove the non-visible vehicles by comparing the distance between the ego-vehicle and the object, i.e. if there is an obstacle, as a building, the real distance will be bigger than the measurement of the depth image, then such “ghost” vehicles will be removed. As we see in the Figure \ref{fig:fig02} this work is not accurate enough. Here we find that some objects are wrongly filtered. The reason for this issue,  is that the distance is dependent on the center of the object and it does not take into account all object dimensions, e.g. such as the center of object is covered by obstacle but the object is partially visible (Figure \ref{equ2}). In addition, this work does not solve the problem of unwanted box, which covers a significant portion of an image. In comparison, our work is checking each pixel in the bounding box, leaves any existing object (even if only partially visible) and filters any object which is completely not visible. The authors \cite{Ref06}
introduced the LGSVL Simulator, which is a high-fidelity simulator designed for developing and testing autonomous driving systems. The simulator provides a comprehensive environment that replicates real-world driving conditions, enabling developers to train and evaluate their autonomous vehicle algorithms under various scenarios.  The LGSVL Simulator is a powerful tool for developing and testing autonomous driving systems. Its high-fidelity representation of the real world, end-to-end simulation flow, customization capabilities, and support for ROS and ROS2 make it an essential tool for researchers, developers, and companies working in the autonomous driving field. In datasets, we could use simulator data or synthetic data, the paper \cite{Ref07} investigates the use of synthetic data to reduce the reliance on real-world data for training object detection models. Object detection is a crucial task in many computer vision applications, including autonomous driving, robotics, and surveillance. However, collecting large amounts of high-quality real-world data can be expensive and time-consuming. Their paper conducts experiment to evaluate the effectiveness of using synthetic data for object detector training. The results show that synthetic data can significantly reduce the requirement for real-world data, with mixed datasets of up to 20\% real-world data achieving comparable detection performance to datasets with 100\% real-world data. Finally, as we are aware there is a data collected by real images, The Canadian Adverse Driving Conditions (CADC) dataset \cite{Ref08} is a publicly available dataset of annotated driving sequences collected in winter conditions in the Waterloo region of Ontario, Canada. It consists of over 20 kilometers of driving data, including over 7,000 annotated frames from eight cameras, a lidar sensor, and a GNSS+INS system. The dataset is annotated with a variety of objects, including vehicles, pedestrians, cyclists, and traffic signs, as well as weather conditions, such as snow, rain, and fog. The CADC dataset is a valuable resource for researchers and developers working on autonomous driving systems. Its large size, diverse collection of driving sequences, high-quality annotations, and multiple sensor modalities make it a powerful tool for improving the performance and robustness of autonomous driving systems in winter conditions.
\section{\uppercase{CONTRIBUTIONS}}\label{sec:CONTRIBUTIONS} 
\begin{figure}[h]
    \centering
    \includegraphics[width=1\linewidth]{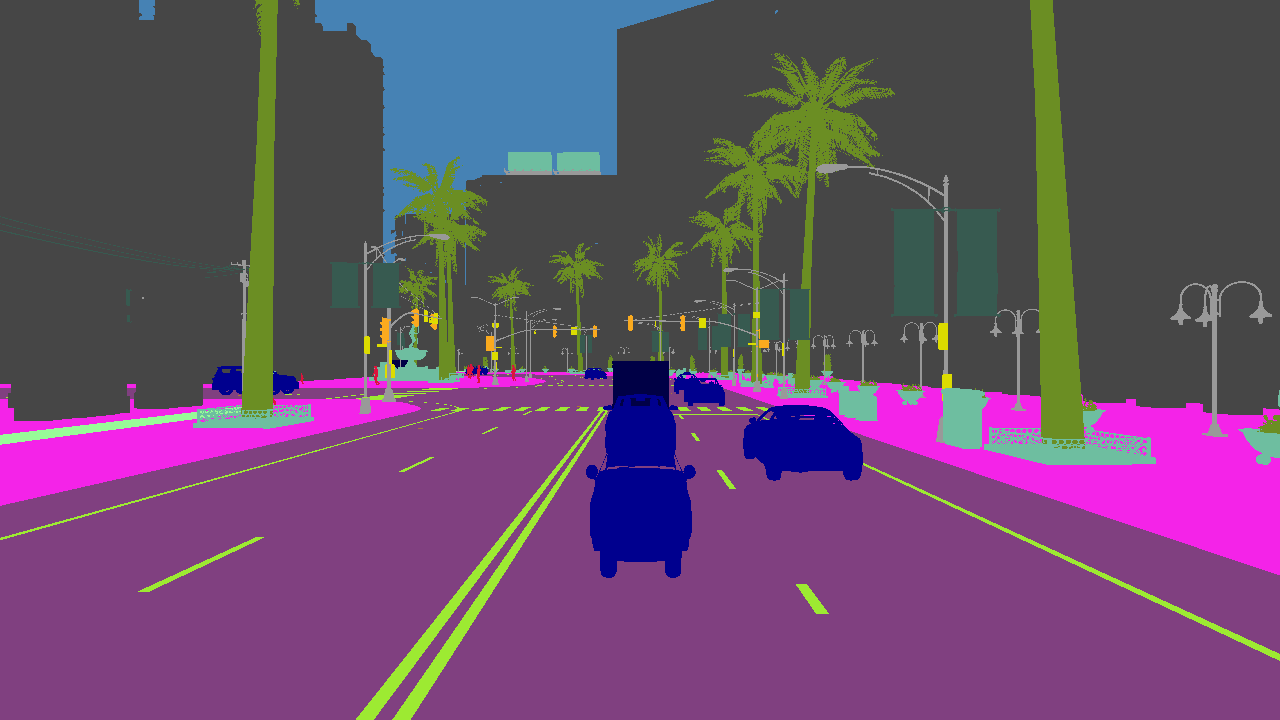}
    \caption{Semantic segmentation camera in Carla Simulator.}
    \label{fig:fig03}
\end{figure}

In this work, we have developed a new technique to collect the necessary data, just by generating it with Carla simulation \cite{Ref09}. Furthermore, we generated a new package to develop the Carla environments where we could control on the number of vehicles, pedestrians, environments, weather, etc by yaml files. The package, developed by us, is called carlasimu, and to simplify the work of other developers, we uploaded it to the GitHub link: (\url{https://github.com/Mofeed-Chaar/Improving-bouning-box-in-Carla-simulator}). The sensors that we have used for our developments in the Carla simulator are RGB Camera sensor and Semantic segmentation camera. You can find our code of the filter at the GitHub link above. As we are aware, the Semantic segmentation camera gives a special pixels colour for each object (Figure \ref{fig:fig03}), in Carla simulator there are 27 classes of objects (Table \ref{tab:tab1}). To improve the bounding boxes, we applied our work to filter the 
“ghost” object, which are occluded by e.g, buildings and which should not be present in the simulation. Simultaneously, we put bounding boxed around existing objects, which are partially covered by other 
objects. The task involves six distinct categories: cars, buses, trucks, vans, pedestrians, and traffic lights. It revolves the “ghost” bounding box around invisible cars, extracting a 2D bounding box, generated by the Carla simulator and conducting a pixel-wise comparison within the box using a semantic segmentation camera. The objective is to determine if the pixels within the box correspond to the object identified by the image Semantic Segmentation sensor. By this operator, we have got highly accurate results by using for training the data which has been collected by our filter (precision = 0. 968, Recall = 0.926, and mAP50 = 0.965) in YOLOv5s. We generated this training data form 8 maps under clear weather conditions.\\
\begin{table}[ht]
\caption{The colour of objects for Semantic segmentation camera in Carla simulator.}
\label{tab:tab1} \centering
\begin{tabular}{|c|c|}
  \hline
  Class Name & Colour (R,G,B) \\
  \hline
  Unlabelled & (0, 0, 0) \\
  \hline
  Car and Truck & (0,0,142) \\
  \hline
  Bus & (0,60,100) \\
  \hline
  Van & (0,0,70) \\
  \hline
  Bicycle & (119,11,32) \\
  \hline
  Motorcycle & (0,0,230) \\
  \hline
  Building & (70, 70, 70) \\
  \hline
  Fence & (100, 40, 40) \\
  \hline
  Other & (55, 90, 80) \\
  \hline
  Pedestrian & (220, 20, 60) \\
  \hline
  Pole & (153, 153, 153) \\
  \hline
  Road Line & (157, 234, 50) \\
  \hline
  Road & (128, 64, 128) \\
  \hline
  Sidewalk & (244, 35, 232) \\
  \hline
  Vegetation & (107, 142, 35) \\
  \hline
  Wall & (102, 102, 156) \\
  \hline
  Traffic Sign & (220, 220, 0) \\
  \hline
  Sky & (70, 130, 180) \\
  \hline
  Ground & (81, 0, 81) \\
  \hline
  Bridge & (150, 100, 100) \\
  \hline
  Rail Track & (230, 150, 140) \\
  \hline
  Guardrail & (180, 165, 180) \\
  \hline
  Traffic Light & (250, 170, 30) \\
  \hline
  Static & (110, 190, 160) \\
  \hline
  Dynamic & (170, 120, 50) \\
  \hline
  Water & (45, 60, 150) \\
  \hline
  Terrain & (145, 170, 100) \\
  \hline
\end{tabular}
\end{table}
\section{\uppercase{BOUNDIG BOX IN CARLA}}\label{sec:BOUNDIG BOX IN CARLA}
\begin{figure}[ht]
    \centering
    \includegraphics[width=1\linewidth]{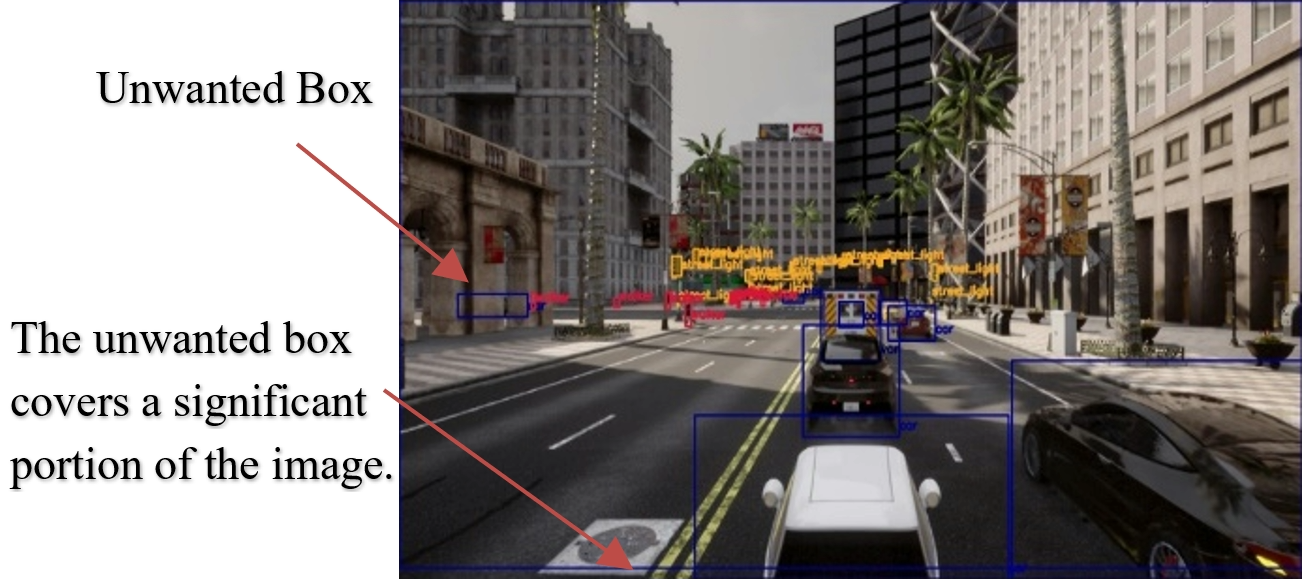}
    \caption{In Carla, as we've found out, objects are detected in bounding box even when they are obscured behind buildings. This affects data generation, where some labels of bounding boxes are false positives, impacting the accuracy of training data.}
    \label{fig:fig04}
\end{figure}

In Carla simulator, the developers use a method which can generate a bounding box (3D, and 2D) for all objects in the environment such as vehicles. For this purpose is used the python class in Carla package (carla.BoundingBox) - see the code in \cite{Ref10}.In this section we will discuss the issues in Auto Bounding box which is generated by Carla simulator.\\ 
The biggest problem appears during the generation of a bounding box in the negative values of the annotation and for objects, which are occluded by obstacles (Figure \ref{fig:fig04}).

\subsection{Negative value in annotations}\label{subsec:Negative value in annotations}
The procedure of the bounding box in the Carla simulator: The bounding box is taken from all the environment, and we choose all the objects in the front of the sensor with respect to the FOV (Field of View) and vertical to the image in the sensor. The issue in this case is that some objects appear partially in the range of the sensor.  This makes a negative value for the bonding box and makes the box appear behind the actual object (as in Figure \ref{fig:fig04}) as a beam for 3D boxes and significant box for 2D bounding box. Thus, it does not fix the “ghost” box problem, when we correct the negative boxes to zero boxes.
The previous works for filtering the bounding box (Section \ref{sec:RELATED WORKS}) did not fix this issue where we operate a special filter for negative notation. Therefore, we could generate a sample of “ghost”-problem typical data in our GitHub link by running the file (boundingBox.py).
\subsection{Ghost Bounding box}\label{subsec:Ghost Bounding box}
The bounding box which is generated by Carla simulator appears for all objects in the coordinate of the view of the camera (FOV and vertical view), and it does not take into account the occluded object by obstacles such as a wall or buildings (Figure \ref{fig:fig04}). The previous works solved this problem, but it is not resolved accurately in some special cases (Section \ref{sec:RELATED WORKS}). Whereas we solved this issue with high accuracy i.e. if an object is present in the image by just one pixel, we can detect it.
\begin{figure}[ht]
    \centering
    \begin{subfigure}{.25\textwidth}
    \centering
    \includegraphics[width=1\linewidth]{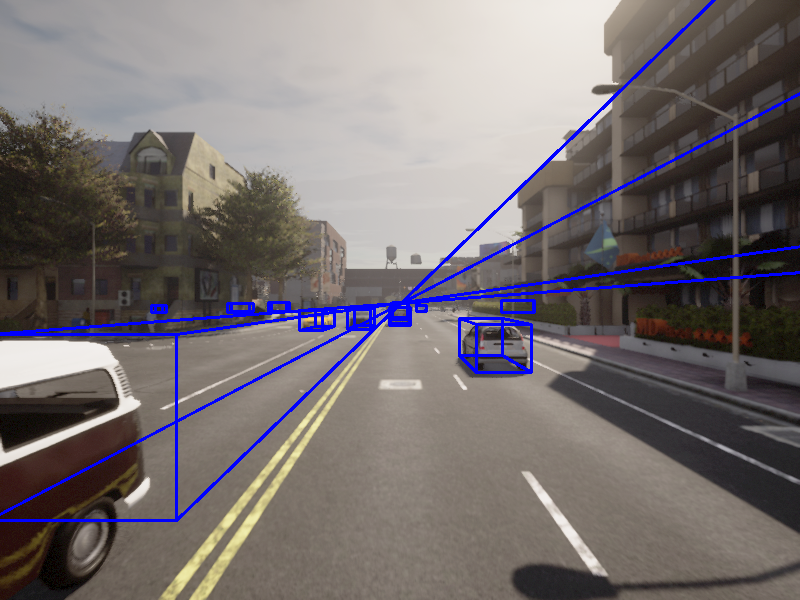}
    \label{fig5:sub1}
    \end{subfigure}%
    \begin{subfigure}{.25\textwidth}
    \centering
    \includegraphics[width=1\linewidth]{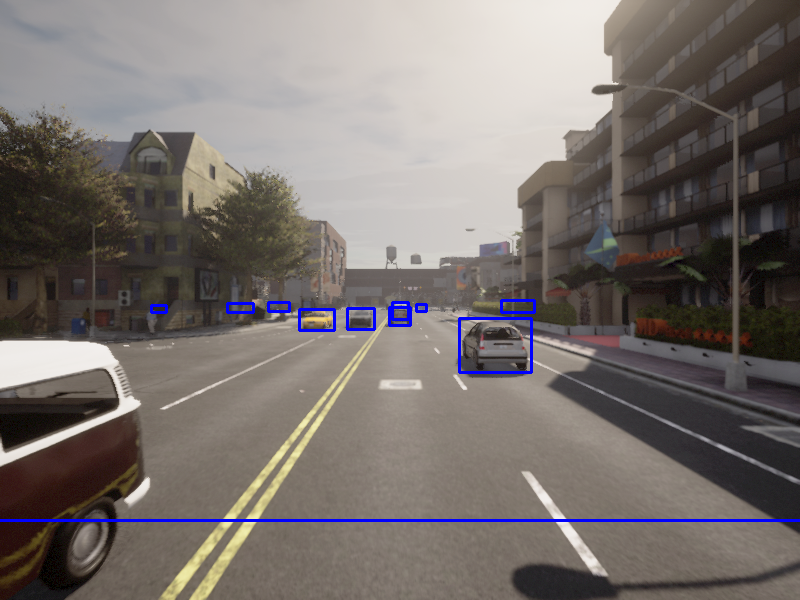}
    \label{fig5:sub2}
    \end{subfigure}
    \caption{The negative notation effects on the bounding boxes for the objects. For 3D bounding box, the box appears as a beam (left). For 2D bounding box (see the right image), the box should have included the vehicle (in the corner of the image at the right) inside, but instead, it is covering a significant portion of the entire image.}
    \label{fig:fig05}
\end{figure}
\section{\uppercase{METHODOLOGIES}}\label{sec:METHODOLOGIES}
\subsection{Objects in Carla simulator}\label{subsec:Objects in Carla simulator}
Carla simulator includes four vehicle classes (car, truck, van, and bus), as well as vulnerable road users (motorcycles, bicycles, and pedestrians). We excluded motorcycles and bicycles from our work because their 2D bounding boxes appear as lines instead of boxes. For static objects, we included traffic lights in our work. Extending the work to include other static objects is possible with our algorithm. The classes we included in our work is (car, bus, truck, van, walker, and traffic light).
\subsection{Sensors in Carla simulator}\label{subsec:Sensors in Carla simulator}
\begin{itemize}
    \item RGB image sensor: This sensor is installed in the ego vehicle in Carla simulator and the output as an RGB (Red, Green, and Blue) image, the parameters of this sensor in our work have the following FOV=90, iso=100, gamma=2.2, and image size= (x:1280, y:720).
    \item Semantic segmentation camera: We also installed Semantic segmentation Camera on the ego vehicle in the simulation.  In the context of a camera, it refers to a computer vision technique used in image analysis and computer vision systems. It involves the process of classifying each pixel in an image into a specific category or class, such as identifying objects, regions, or areas in the image and labelling them accordingly. This technique is commonly used in various applications, including autonomous vehicles, surveillance, image editing, and medical imaging, among others. Applications of semantic segmentation with cameras include Autonomous Vehicles \cite{Ref11} where Semantic segmentation is crucial for self-driving cars to identify and understand their surroundings, including detecting pedestrians, other vehicles, road signs, and road boundaries. In Carla simulator we used automatic labelling of the object (Table \ref{tab:tab1}) which plays a major role in our work, and we take the same parameters as the RGB camera sensor (ROV and Image Size) \cite{Ref12} 
    \item Radar: The RADAR sensor in the CARLA simulator is a placeholder model that is not based on raytracing. It casts rays to objects and computes distance and velocity using a simplified model. The sensor can detect obstacles in front of it and can be used for applications such as adaptive cruise control and obstacle detection. The output of radar is altitude, azimuth, depth, velocity.
    \item Lidar: The CARLA LiDAR sensor is a 3D point cloud sensor that uses laser light to measure distance to objects in its surroundings.
    \item Depth image: CARLA depth images are grayscale images where each pixel represents the distance to the object at that pixel. There are two kinds of depth image: colour image and gray scale image, which turned the distance sorted as an RGB channels into a [0,1] float. In our generated Data, we generated the Depth image as grayscale.
\end{itemize}
Note: The parameters in this sensors are editable where we could control in the parameters in the file (sensors.yaml) – see the package in our GitHub project, where we generated RGB images, Semantic segmentation images, radar data, lidar data, depth images, and filtered bounding box notations as an text file.
\subsection{Correct the coordinates of the boxing object}\label{subsec:Correct the coordinates of the boxing object}
\begin{figure}[ht]
    \centering
    \includegraphics[width=1\linewidth]{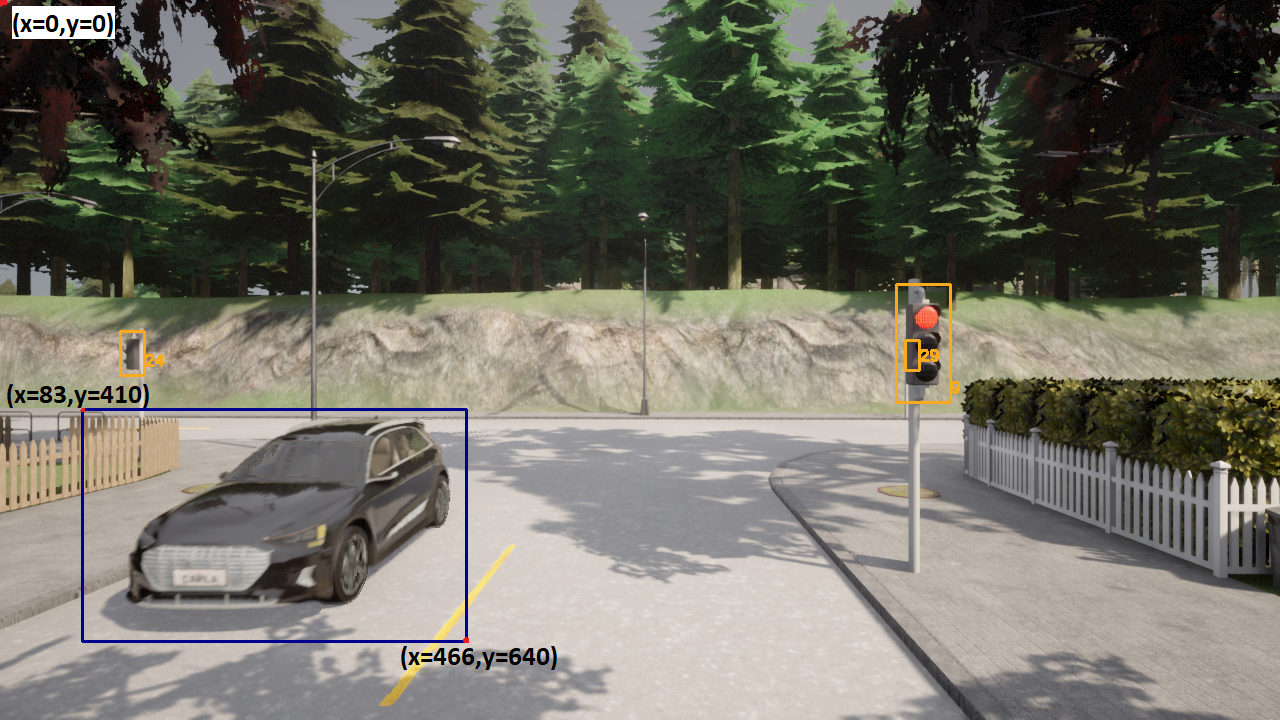}
    \caption{The coordinates of the box of an object are shown. We defined the corner coordinates of the box on the car as (x min, y min, x max, y max) = (83, 410, 466, 640) respectively.}
    \label{fig:fig06}
\end{figure}

\begin{figure}[ht]
    \centering
    \includegraphics[width=1\linewidth]{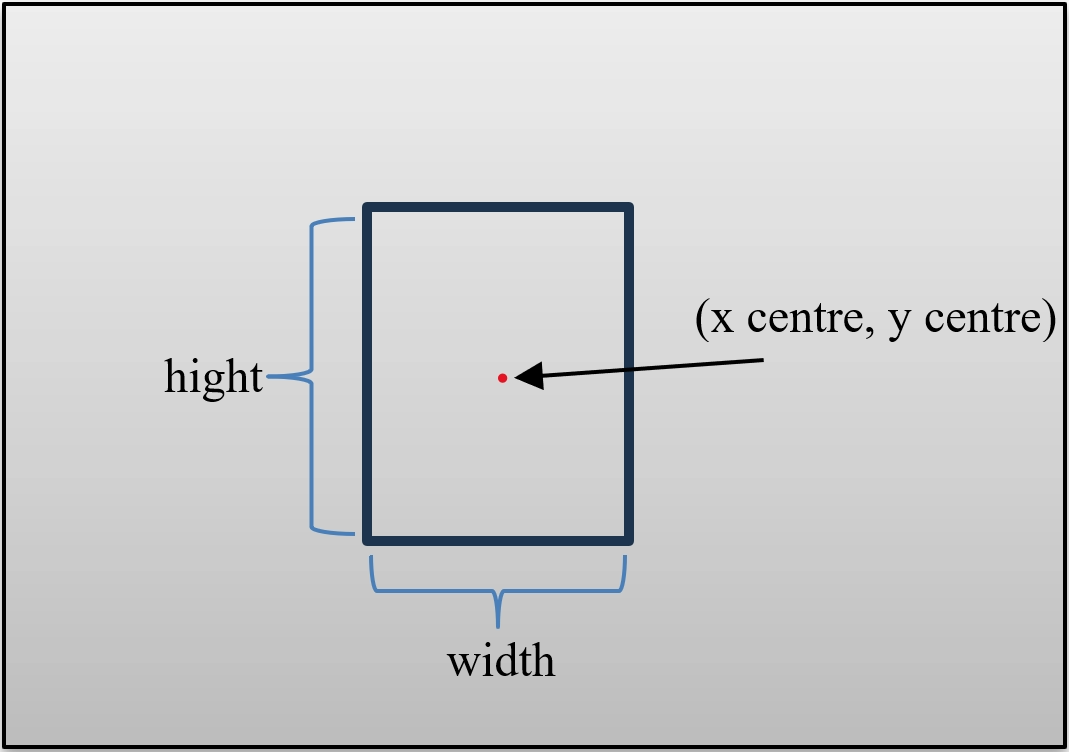}
    \caption{Centre coordinates for bonding box objects}
    \label{fig:fig07}
\end{figure}

The 2D box notation is defined in two ways: the centre box and the corner box. The corner box is represented as (x min, y min, x max, y max), with the reference point located at the upper-left corner of the image. The numerical values within this notation correspond to the coordinates of the pixels at that specific point (Figure \ref{fig:fig06}). In addition, the coordinates should be normalized \cite{Ref13} according to the following formula:
\begin{equation}\label{eq3}
\begin{aligned}
    x_{min} = x_{up\_ left}/width\\
    y_{min} = y_{up\_ left}/hight\\
    x_{max} = x_{down\_ right}/width\\
    y_{max} = y_{down\_ right}/hight
\end{aligned}
\end{equation}
Where the width and hight describe the dimension of image by pixels. The centre coordinates describe the centre, width and hight of the box (x centre, y centre, width, hight) (Figure \ref{fig:fig07}).
It's important to note that the minimum value for a coordinate is always zero. However, the maximum value varies depending on whether the coordinate has been normalized or not. In the case of normalized coordinates, the maximum value is set to one. On the other hand, when dealing with non-normalized coordinates, the upper limit is determined by the dimensions of the image pixel. It's worth mentioning that in the context of the coordinate system used in the Carla simulator, it's possible for a generated corner coordinate to fall below zero or exceed the dimensions of the image pixel. To correct this issue, we filtered the value of coordinates as follows:
\begin{algorithm}[!h]
 \caption{Correct the coordinate.}
 \KwData{Normalized corner box (x centre, y centre, width, hight)}
 \KwResult{Correct the coordinate of the box }
  \If{\(x_{min}>1\)}{
   delete the box and stop\;
   }
  \(x_{min}=max(0,x_{min})\)\;
  \If{\(x_{max}<0\)}{
   delete the box and stop\;
   }
   \(x_{max}=min(x_{max},1)\)\;
   Repeat same steps for \(y_{min}\) and \(y_{max}\)\;
   \If{area of the box = 0 }{
   delete the box\;
   }
\end{algorithm}
\FloatBarrier
\noindent The correction of this issue of coordinates will resolve the incorrect bounding box value. However, it's important to note that this correction won't address the box covering a significant portion of the image (Figure \ref{fig:fig04}). We will delve into how we addressed this challenge and how we did solve it later in this paper.
\subsection{Filter unwanted boxes}\label{subsec:Filter unwanted boxes}
\begin{figure}[ht]
    \centering
    \includegraphics[width=1\linewidth]{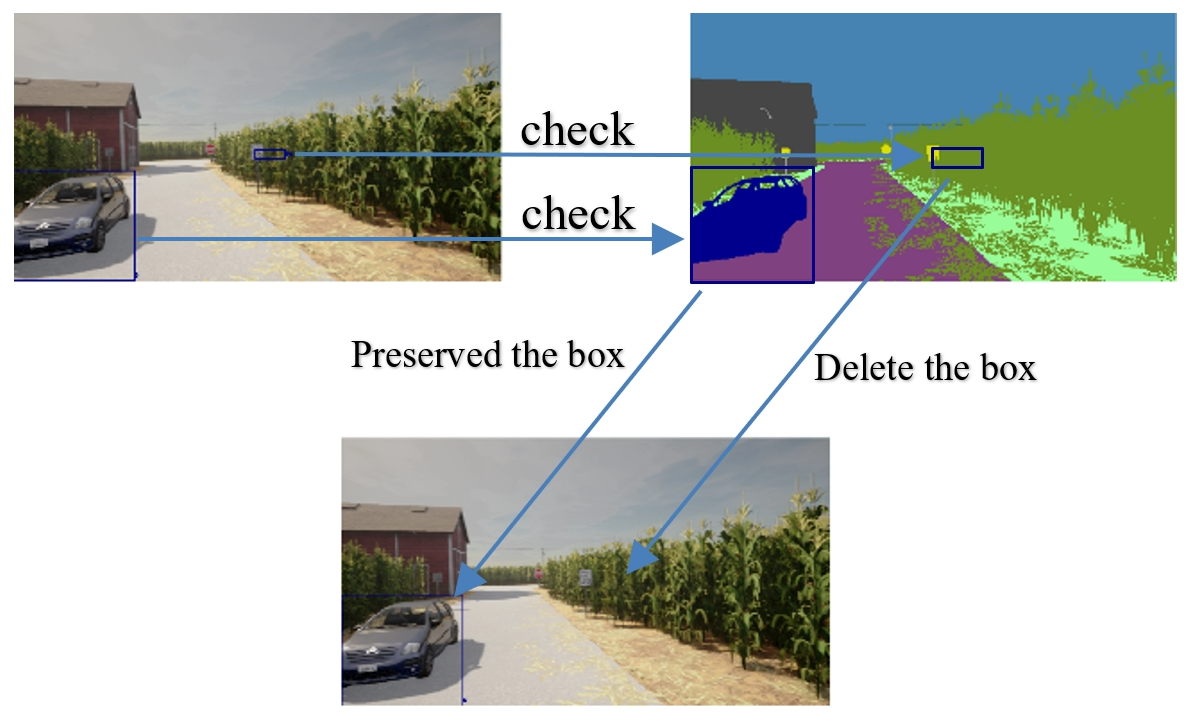}
    \caption{In this filter, we check each box in image segmentation camera if it fulfils 10\% of the colour corresponding to the object in image segmentation. In our example the colour of the car in image segmentation in Carla is (0, 0, 142), the small box behind the field is hidden and there is no pixel with value (0, 0, 142) inside of it, as compared to the bounding box for the car in the large box.}
    \label{fig:fig08}
\end{figure}

As discussed earlier, another issue related to bounding boxes in Carla simulator is the presence of "ghost boxes." These are boxes that are occluded by other objects (and not visible in the simulation) but they are still labelled in Carla (Figure \ref{fig:fig01}). To address this challenge, we utilize semantic segmentation in the Carla simulator to filter out unwanted boxes. In the Carla simulator, there are 28 distinct colours associated with various objects (as shown in Table \ref{tab:tab1}). Our approach involves systematically examining each bounding box using semantic segmentation cameras. We verify whether the colour within the box corresponds to the expected object colour. For instance, if the box represents a car, we check if there are pixels with the colour (0,0,142) inside the box, aligning with the designated area in the image segmentation camera. To enhance the precision of the bounding box, we implemented a filtering criterion to refine object detection. Specifically, we retained bounding boxes that occupied at least 10\% of the total area. In other words, if at least 10\% of the bounding box of the car area contained pixels with a value of (0, 0, 142) (Table \ref{tab:tab1}), the box was preserved, otherwise, it was deleted (Figure \ref{fig:fig08})
The threshold of 10\% is a flexible value in our work. We can change the boundingBox.yaml file in the GitHub project by changing the threshold\_small\_box to the appropriate present. This allows us to adjust the size of the bounding boxes that are detecting. For example, if we want to detect smaller objects, we can lower the threshold value. The threshold\_small\_box parameter is located in the boundingBox.yaml file. We can change the value of this parameter by editing the file and saving it. The new value will be used the next time we run the object detection code. Here is an example of how to change the threshold\_small\_box parameter:
\begin{small}
\begin{verbatim}
            threshold_small_box: 5
\end{verbatim}
\end{small}
This will change the threshold value to 5\%. This means that only bounding boxes, that are at least 5\% of the area, which contains the colour corresponding to this object on image segmentation, will be detected. This procedure allows us to check whether the object exists at each pixel of the bounding box, so we can detect the object even if the box has only one pixel.
\subsection{Filter the large boxes}\label{subsec:Filter the large boxes}
\begin{figure}[ht]
    \centering
    \includegraphics[width=1\linewidth]{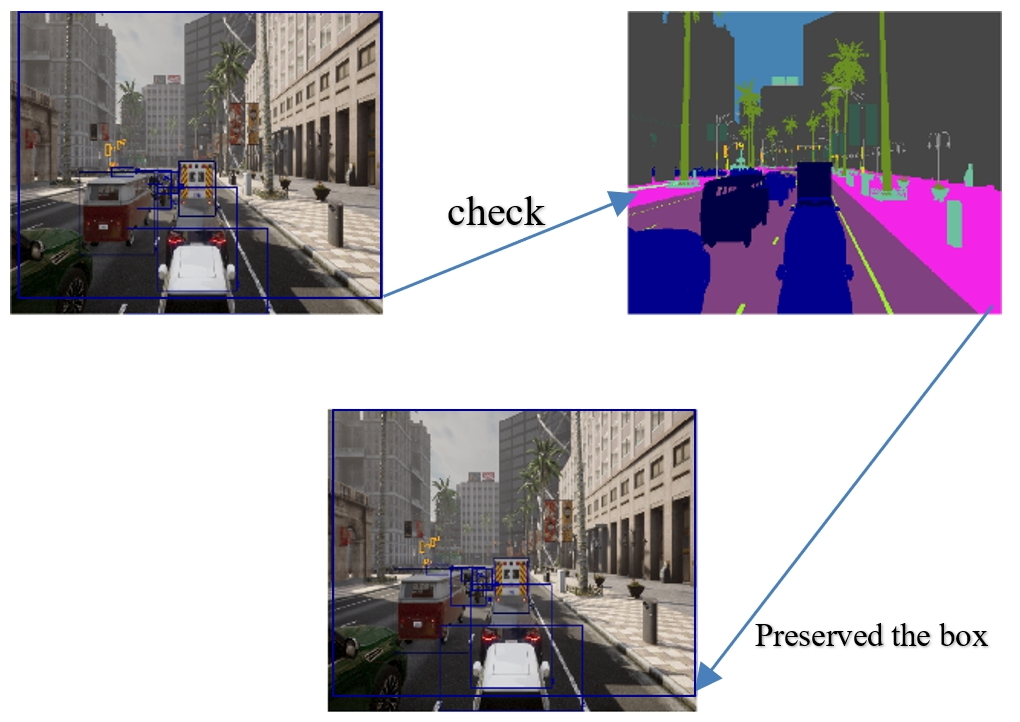}
    \caption{The large box that covers a significant portion of the image and contains multiple cars that exceed the threshold (10\%) of the area of the box, is a ghost box for a car. This leads to preserving the bounding box, even though it does not accurately identify a car.}
    \label{fig:fig09}
\end{figure}

The previous method (Section \ref{subsec:Filter unwanted boxes}) suffers one limitation in the case of large bounding boxes. This is because a bounding box that identifies an object may contain other objects, if the bounding box is large enough to cover them. Therefore, if we use a 10\% match threshold, then the bounding box may be incorrectly retained, even if it does not accurately identify the object (Figure \ref{fig:fig09}).
\\
To solve this challenge, we defined two thresholds for the large boxes.
\begin{itemize}
\item The ratio of the bounding box area to the image area.
\item A special threshold for filtering large bounding boxes.
\end{itemize}
In our work, we used a criterion of 70\% to determine whether a bounding box is large. In other words, if the ratio of the bounding box area to the image area is greater than 70\%, then we considered the box to be large. Otherwise, the box was considered to be normal or small. The special threshold that we used in our GitHub project to filter the boxes is 50\%. This means that if a large bounding box contains more than 50\% of pixels in the image segmentation that belong to the same class as the detected object, then the box is preserved. Otherwise, the box is deleted. This algorithm describes the filtering of the box:
\begin{algorithm}[!h]
 \caption{Filter the boxes.}
 \KwData{Normalized corner box (x centre, y centre, width, hight)}
 \KwResult{Filter the ghost boxes.}
 initialization\;
 Count the pixels = 0\;
 \For{all the pixels in the box in image segmentation}{
   \If{the pixel is the same class of the box detected }{
   Count the pixels +=1\;
   }
   }
  \eIf{the box is large}{
  \If{the ratio of Count the pixels to the number of pixels of the box is less than the large box threshold filter}{
  delete the box\;
  }
  }{
  \If{the ratio of Count the pixels to the number of pixels of the box is less than the normal box threshold filter}{
  delete the box\;
  }
  }
 \end{algorithm}
\subsection{Filter 3D bounding box}\label{subsec:Filter 3D bounding box}
\begin{figure}[ht]
    \centering
    \includegraphics[width=1\linewidth]{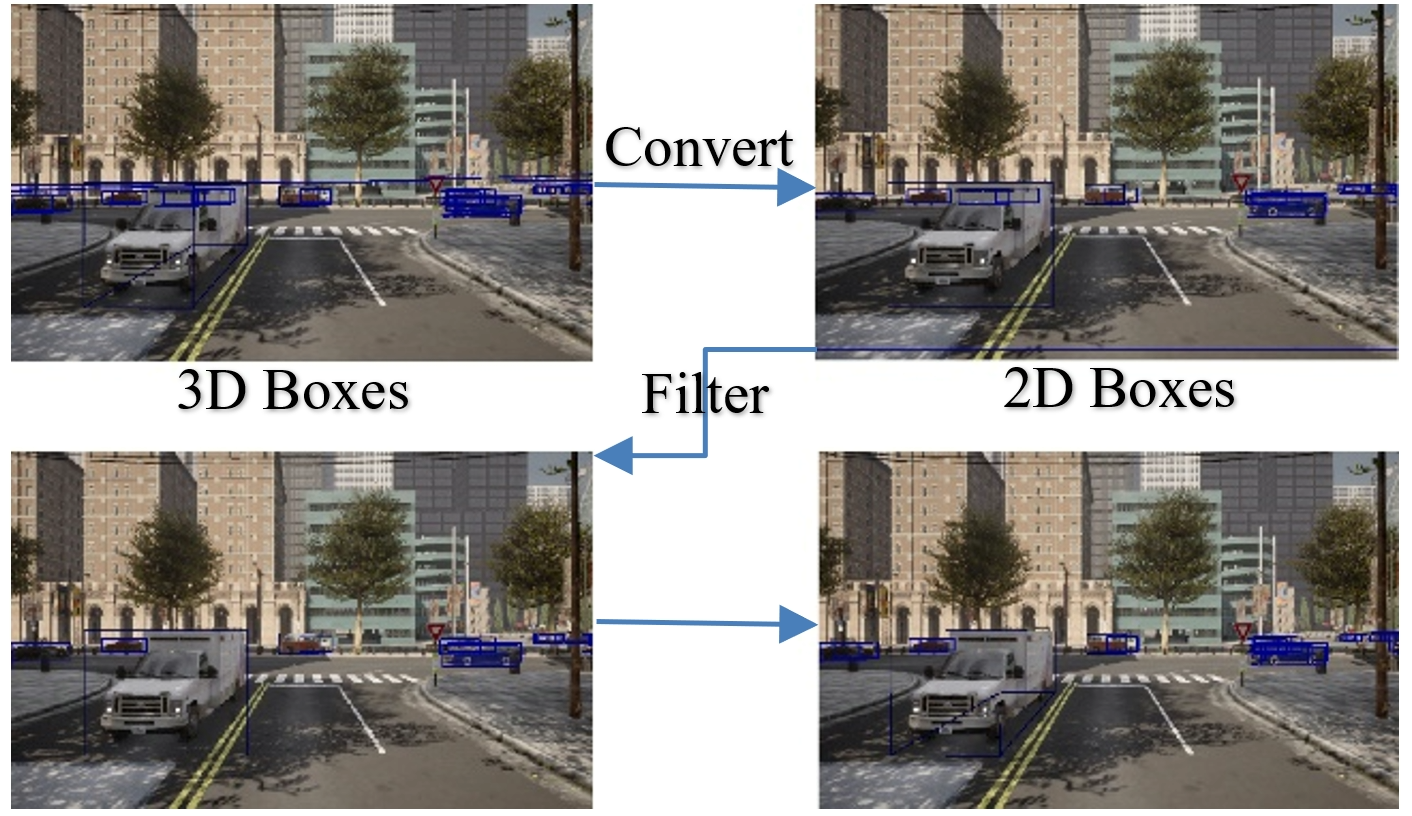}
    \caption{Filtering the 3D bounding box by check if the box is ghost box or not in 2D bounding box.}
    \label{fig:fig10}
\end{figure}

In our work, we propose a novel filtering approach for 3D bounding boxes by converting the boxes to 2D bounding boxes and applying a filtering criterion to the 2D representations. This approach enables efficient and effective filtering of 3D bounding boxes (Figure \ref{fig:fig10}). Filtering 3D bounding boxes by identifying ghost boxes, which are bounding boxes that do not correspond to real objects. This is a crucial step in object detection and tracking tasks. In our work, we propose a novel filtering method that leverages the concept of 2D bounding boxes to efficiently identify and remove ghost boxes from 3D bounding box sets. By converting 3D bounding boxes to their corresponding 2D representations, we can effectively check for ghost boxes based on 2D geometric criteria. This approach significantly improves the accuracy and efficiency of object detection and tracking tasks, particularly in scenes with complex backgrounds or occlusions. In the GitHub we implemented an example to filter the 3D bounding box in the file name 3D\_Bounding\_Box.py.
\section{OUR DATA VALIDATION}\label{sec:OUR DATA VALIDATION}
\begin{figure}[ht]
    \centering
    \includegraphics[width=1\linewidth]{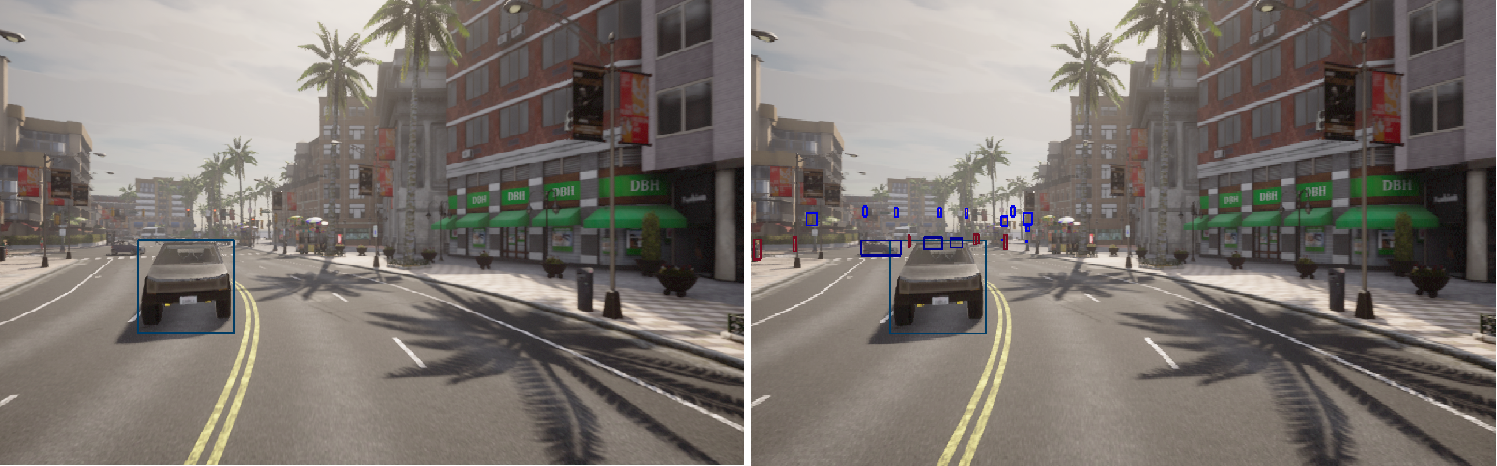}
    \caption{labelling of bounding boxes up to 50m (left) and up to 100 m (right).}
    \label{fig:fig11}
\end{figure}

\begin{figure}[ht]
    \centering
    \includegraphics[width=1\linewidth]{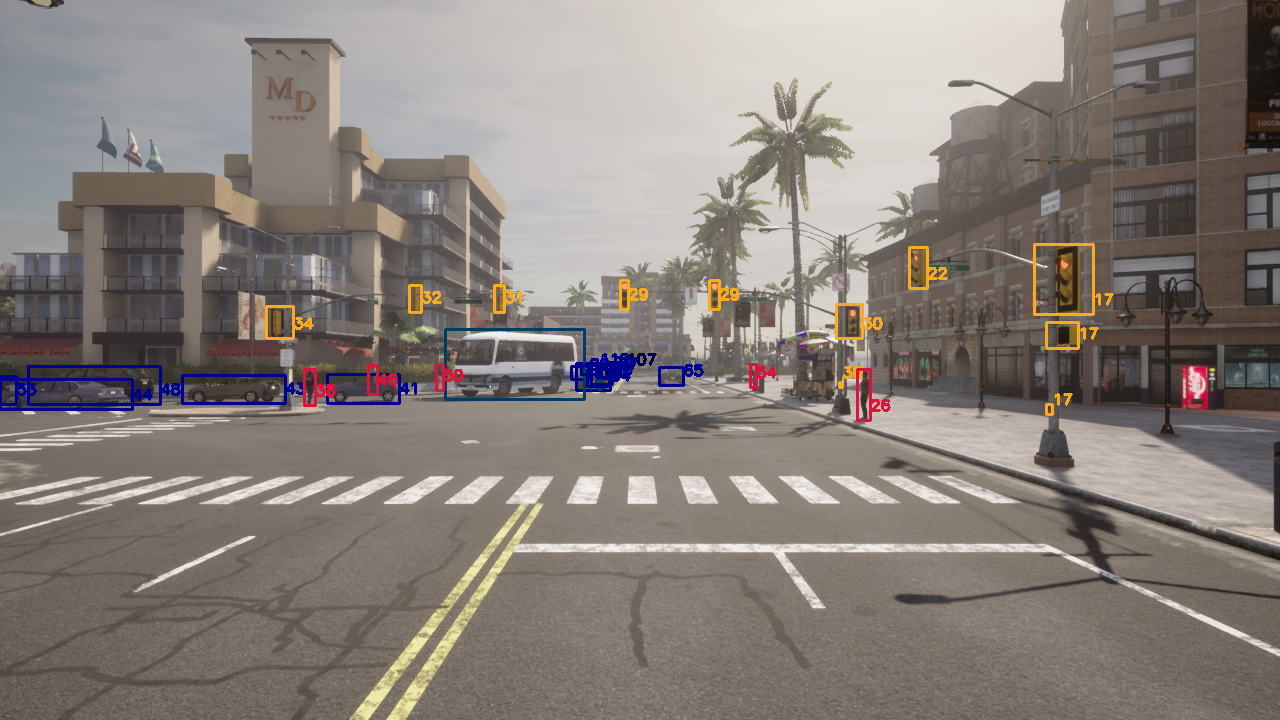}
    \caption{Sample of our data where we see that the traffic light has figures such as a traffic light poles.}
    \label{fig:fig12}
\end{figure}

To create a diverse and comprehensive dataset for object detection and localization, we employed the Carla simulator 9.14 \cite{Ref01} and utilized eight distinct maps (5000 images for each map). Our filtering approach yielded a dataset encompassing six object classes: car, bus, truck, van, walker, and traffic light (see section  \ref{subsec:Filter unwanted boxes}). To enhance the dataset's variety, we incorporated data from five sensors: RGB camera, semantic segmentation camera, depth camera, radar sensor, and LIDAR sensor. For all objects within a range of 50 meters, 100 meters, 150 meters, and 200 meters (Figure \ref{fig:fig11}), we applied our bounding box filter to the RGB images (1280,720). To standardize our results, we trained our dataset using a pretrained YOLOv8 model \cite{Ref14} for bounding boxes with all objects up to 100 meters. The training parameters included SGD (lr=0.01, momentum=0.9), image size for training 640, epoch 50, and iou 0.7 Using 20\% of images for validation (i.e. 1000 images for each map) we got a result for data labelled up to 100 meter as the in Table \ref{tab:tab2}, and results for the same data labelled up to 50 m (Table \ref{tab:tab3}).
\begin{table}[ht]
\caption{A summary of results from the model which is trained using our filtered data in base of bounding box, which is labeling all objects up to 100 meters, where the image size for training has 640x640 pixels.}
\label{tab:tab2} 
\centering
\begin{tabular}{|c|c|c|c|}
  \hline
  Class & Precision & Recall & mAP50 \\
  \hline
  All & 0.915 & 0.664 & 0.75 \\
  \hline
  Car & 0.895 & 0.788 & 0.859 \\
  \hline
  Bus & 0.925 & 0.949 & 0.961 \\
  \hline
  Truck & 0.909 & 0.433 & 0.536 \\
  \hline
  Van & 0.874 & 0.795 & 0.865 \\
  \hline
  Walker & 0.914 & 0.485 & 0.585 \\
  \hline
  Traffic light & 0.972 & 0.535 & 0.696 \\
  \hline
\end{tabular}
\end{table}
\begin{table}[ht]
\caption{A summary of results from the model which is trained using our filtered data in base of bounding box, which is labeling all objects up to 50 meters, where the image size for training has 640x640 pixels.}
\label{tab:tab3} 
\centering
\begin{tabular}{|c|c|c|c|}
  \hline
  Class & Precision & Recall & mAP50 \\
  \hline
  All & 0.947 & 0.805 & 0.882 \\
  \hline
  Car & 0.941 & 0.908 & 0.964 \\
  \hline
  Bus & 0.917 & 0.981 & 0.983 \\
  \hline
  Truck & 0.965 & 0.524 & 0.639 \\
  \hline
  Van & 0.927 & 0.918 & 0.965 \\
  \hline
  Walker & 0.954 & 0.757 & 0.865 \\
  \hline
  Traffic light & 0.977 & 0.739 & 0.876 \\
  \hline
\end{tabular}
\end{table}
\FloatBarrier \noindent Form the previous two tables, we can conclude that the accuracy of object detection decreases as the distance between the object and the camera increases. This is because smaller objects are more difficult to detect due to their lower resolution. This effect is particularly pronounced when annotating objects for distant scenes. To improve the accuracy of object detection for distant objects, it is important to use high-resolution images and to train the object detection model on a dataset that includes a large number of distant objects.
\begin{table}[ht]
\caption{A summary of results from the model which is trained using our filtered data in base of bounding box, which is labeling all objects up to 100 meters, where the image size for training has 1280x1280 pixels.}
\label{tab:tab4} 
\centering
\begin{tabular}{|c|c|c|c|}
  \hline
  Class & Precision & Recall & mAP50 \\
  \hline
  All & 0.939 & 0.787 & 0.876 \\
  \hline
  Car & 0.927 & 0.852 & 0.919 \\
  \hline
  Bus & 0.918 & 0.953 & 0.977 \\
  \hline
  Truck & 0.969 & 0.615 & 0.787 \\
  \hline
  Van & 0.906 & 0.866 & 0.924 \\
  \hline
  Walker & 0.944 & 0.687 & 0.8 \\
  \hline
  Traffic light & 0.971 & 0.75 & 0.851 \\
  \hline
\end{tabular}
\end{table}
\begin{table}[ht]
\caption{A summary of results from the model which is trained using our filtered data in base of bounding box, which is labeling all objects up to 50 meters, where the image size for training has 1280x1280 pixels.}
\label{tab:tab5} 
\centering
\begin{tabular}{|c|c|c|c|}
  \hline
  Class & Precision & Recall & mAP50 \\
  \hline
  All & 0.955 & 0.908 & 0.966 \\
  \hline
  Car & 0.962 & 0.932 & 0.979 \\
  \hline
  Bus & 0.916 & 0.982 & 0.989 \\
  \hline
  Truck & 0.948 & 0.734 & 0.909 \\
  \hline
  Van & 0.949 & 0.967 & 0.987 \\
  \hline
  Walker & 0.975 & 0.895 & 0.961 \\
  \hline
  Traffic light & 0.98 & 0.935 & 0.97 \\
  \hline
\end{tabular}
\end{table}
\FloatBarrier \noindent As evident in Table \ref{tab:tab4} and Table \ref{tab:tab5}, training our data using an image size of 1280 resulted in enhanced object detection accuracy, particularly for objects located at greater distances. This improvement can be attributed to the increased resolution of the images, which allows the object detection model to capture finer details and distinguish objects more effectively, especially when dealing with smaller objects at a distance. Interestingly, we observed that traffic lights, despite being small objects, achieved high detection accuracy. This can be attributed to the distinctive shape of traffic lights, which resembles traffic light poles (Figure \ref{fig:fig12}). The object detection model is likely to learn these distinctive features, enabling it to accurately identify traffic lights even when they are small or hight distant.
\section{CONCLUSIONS}\label{CONCLUSIONS}
The Carla simulator is an effective tool for generating datasets for object detection tasks. Its flexibility and controllability over the environment and artificial scenarios, such as accidents, congestion, and severe weather conditions, make it a valuable tool for creating realistic and challenging datasets. Our proposed filter has been shown to be highly effective in improving the accuracy of object detection models trained by Carla datasets generated using our filter. We believe that our work represents a significant step forward in the development of high-quality datasets for object detection tasks. In addition, data generation through the CARLA simulator has become more reliable. The project we developed to create bounding boxes is now fully available on GitHub. In addition, we have made it more flexible to select parameters in CARLA through YAML files. This allows for the generation of data related to weather conditions, such as fog, rain, and the number of cars, as well as the ability to control car lights. This flexibility makes it easy to develop self-driving cars in severe weather conditions in CARLA. In addition, we have included many sensors, such as radar, lidar, and depth image, which allow for the integration and synchronization of multiple sensors and the use of the bounding boxes that we developed. Using YOLOv5 and YOLOv8, we achieved good results and high accuracy in the data that was collected using our algorithm to filter bounding boxes. The accuracy exceeded 90\%. This confirms the success of the filter we developed in generating data through the CARLA simulation program.

\bibliographystyle{apalike}
{\small
\bibliography{bibliography}}
\end{document}